\documentclass[twoside,11pt]{article}
\usepackage{jmlr2e}

\usepackage[utf8]{inputenc}
\usepackage[T1]{fontenc}
\usepackage{url}
\usepackage{booktabs}
\usepackage{amsfonts}
\usepackage{nicefrac}
\usepackage{microtype}
\usepackage{xcolor}
\usepackage{xspace}
\usepackage{graphicx}
\usepackage{amssymb,amsmath}
\usepackage{algorithm}
\usepackage{algorithmic}
\usepackage{paralist}
\usepackage{multirow}
\usepackage{enumerate,enumitem}
\usepackage{makeidx}
\usepackage{mathtools}
\usepackage{epstopdf}
\usepackage{mathrsfs}
\usepackage{bm}
\usepackage{bbm}
\usepackage{upgreek}
\usepackage{cleveref}
\usepackage{dsfont}

\newcommand{\ben}{\begin{enumerate}}
\newcommand{\een}{\end{enumerate}}

\newcommand{\cmt}[1]{}

\newcommand{\cD}{\mathcal{D}}

\title{Verbalized Particle Posterior: Bayesian Inference over\\
Natural Language Hypotheses}

\author{%
\name Yan Zhang \email yz18b@fsu.edu \\
\addr Department of Computer Science, Florida State University
\AND
\name Shikan Lian \email shikanlian@gmail.com
\AND
\name Shibo Li\thanks{Corresponding author.} \email shiboli@cs.fsu.edu \\
\addr Department of Computer Science, Florida State University
}

\begin{document}
\firstpageno{1}

\begingroup
\renewcommand\thefootnote{\fnsymbol{footnote}}
\maketitle
\endgroup

\begin{abstract}
Verbalized Machine Learning (VML) parameterizes a model as a natural-language prompt that an LLM evaluates as $f(x; \theta)$. The framework is interpretable, but it commits to a single hypothesis with no measure of uncertainty, and that hypothesis varies substantially across optimization runs on the same data. We propose the Verbalized Particle Posterior (VPP), which treats verbalized learning as a Bayesian inference problem: maintain a population of natural-language hypotheses as particles, update them with Metropolis--Hastings (VPP-MH) or Sequential Monte Carlo (VPP-SMC), and predict by Bayesian model averaging. Both algorithms treat the LLM as a black box, requiring no access to logits or gradients. A distinctive consequence follows. In classical Bayesian learning, model selection sits outside the posterior; in VPP both model structure and parameters share a single language space, and the posterior ranges over both. We evaluate VPP on regression, classification, and rule-discovery benchmarks. It improves over a single VML run on every benchmark and matches or exceeds an oracle-best ensemble of independent VML runs on most, while eliminating the catastrophic single-run failures that VML occasionally produces. Because each particle is a human-readable hypothesis, the posterior is itself something a reader can inspect, seeing in plain text which explanations the data supported and which it ruled out.
\end{abstract}

\section{Introduction}
\label{sec:intro}

Verbalized Machine Learning (VML)~\cite{xiao2025vml} parameterizes machine learning models entirely in natural language. An LLM with a text prompt $\theta$ acts as a function $f(x; \theta)$, where $\theta \in \Theta_{\text{language}}$ is constrained to be human-readable text, and a separate optimizer LLM iteratively refines $\theta$ by examining mini-batch predictions against ground-truth labels. The framework is appealing for three reasons: it discovers the appropriate model class from data without the practitioner specifying linear, periodic, or piecewise; it produces hypotheses a human can read directly; and it accepts prior knowledge written in plain language.

These properties come at a cost. VML returns a single hypothesis $\theta^*$ with no notion of uncertainty, and the optimizer's stochasticity over a large discrete space means the returned hypothesis can vary by orders of magnitude across runs. Across $15$ independent VML runs on linear regression in our experiments, test MSE ranges from $1.08$ to $48.30$, a $44.7\times$ spread that gives the practitioner no way to tell whether a particular run was a good or bad draw. The variance is not cosmetic: VML accepts every optimizer proposal without filtering, so a single poor revision can overwrite a good hypothesis with no path back, and individual runs on contains-zero range from $32\%$ to $100\%$ accuracy on the same dataset. And even when a run does fail, VML cannot signal it---there is no mechanism to flag inputs on which its hypothesis is unreliable.

A posterior $p(\theta \mid \mathcal{D})$ is the standard remedy: rather than one hypothesis the data weakly supports, maintain a distribution over many. Putting Bayes on VML is unusual on both sides. The parameter space $\Theta_{\text{language}}$ is discrete, compositional, and admits no gradients or continuous neighborhoods, so most modern posterior-inference machinery is unavailable. The other side is more interesting. In classical Bayesian learning the practitioner first chooses a model family---linear regression, Gaussian process, neural network---and only then performs posterior inference within that family; model selection sits outside the framework. In VML, model structure and parameters live in the same language space, so a single posterior over verbalized hypotheses simultaneously selects the model and infers its parameters. Classical MCMC has no natural way to express that target.

We propose the \emph{Verbalized Particle Posterior} (VPP), a particle-based framework for Bayesian inference over verbalized hypotheses. VPP maintains $K$ natural-language hypotheses as particles, updates them via inference algorithms that treat the LLM as a black box, and predicts by Bayesian model averaging. We instantiate VPP two ways. \textbf{VPP-MH} runs Metropolis--Hastings on each particle, using the optimizer LLM as an informed proposal and accepting each proposal with probability set by the batch likelihood ratio, so updates that would lower the likelihood are likely to be rejected---directly addressing the failure mode in which a single bad revision overwrites a good hypothesis. \textbf{VPP-SMC} additionally maintains importance weights on the particles, reweights them by accumulated likelihood, and resamples when the effective sample size drops, concentrating the ensemble on well-supported hypotheses while letting weak ones be replaced. Because every particle is a complete verbal hypothesis, the resulting posterior is inspectable: a reader can see which hypotheses receive high weight and which the data has ruled out.

The pattern-discovery task \emph{sum-parity}---in which the model must discover from labeled examples that the label is $1$ iff the digit sum is even---illustrates why posterior reweighting differs from extra compute. VPP-SMC reaches $100\%$ test accuracy on every seed. With five times the budget, an oracle that picks the best of five independent VML runs also reaches $100\%$; but the corresponding unweighted ensemble averages only $67.2\%$, because incorrect runs outvote correct ones. VPP-SMC stays perfect because likelihood-based reweighting concentrates weight on the rule the data supports.

\paragraph{Contributions.}
\textbf{(i)} We formulate posterior inference over verbalized hypotheses and instantiate it with two black-box algorithms, VPP-MH and VPP-SMC.
\textbf{(ii)} On seven benchmarks spanning regression, classification, and rule discovery, VPP improves over single-run VML on every task and matches or exceeds an oracle-best $5\times$ ensemble of independent VML runs on six of seven, reducing test MSE by up to $24\times$ on polynomial regression.
\textbf{(iii)} VPP turns particle disagreement into an interpretable uncertainty signal, a capability standard VML lacks.
\textbf{(iv)} We compare MH and SMC in this setting and ablate the key design choices: the Metropolis--Hastings accept/reject step (worth $22.2$\,pp on sum-parity) and the particle count $K$.

\section{Background}
\label{sec:background}

\paragraph{Verbalized Machine Learning.} VML~\cite{xiao2025vml} treats a text prompt $\theta \in \Theta_{\text{language}}$ as the parameter of a model, with an LLM realizing the function $f(x; \theta)$. A separate optimizer LLM refines $\theta$ from mini-batches of (input, prediction, target) triples to minimize the empirical loss
\begin{equation}
	\min_{\theta \in \Theta_{\text{language}}} \frac{1}{N} \sum_{n=1}^N \ell\big(y_n,\, f(x_n; \theta)\big),
	\label{eq:vml_obj}
\end{equation}
where $\ell$ is squared error for regression or zero-one loss for classification. The initial prompt $\theta_0$ is deliberately model-class-agnostic (e.g., ``\textit{You are designed to do regression}''), so the optimizer must induce both functional form and parameters from data. Standard VML returns a single point estimate $\theta^*$---the prompt that survived the last optimizer step---and uses it directly for prediction, with no mechanism for combining or comparing the many alternative hypotheses the optimizer rejected along the way.

\paragraph{Bayesian inference.} Given data $\cD = \{(x_n, y_n)\}_{n=1}^N$ and parameter $\theta$, Bayesian inference replaces the point estimate with the posterior
\begin{equation}
	p(\theta \mid \cD) \;=\; \frac{p(\cD \mid \theta)\, p(\theta)}{p(\cD)} \;\propto\; p(\cD \mid \theta)\, p(\theta),
	\label{eq:bayes}
\end{equation}
which assigns a probability to every $\theta$ in proportion to how well it explains the data, scaled by the prior $p(\theta)$. Predictions at a new $x^*$ are obtained by averaging the model over the posterior,
\begin{equation}
	p(y^* \mid x^*, \cD) \;=\; \int p(y^* \mid x^*, \theta)\, p(\theta \mid \cD)\, d\theta,
	\label{eq:predictive}
\end{equation}
so the prediction reflects the spread of plausible explanations rather than a single committed one. The integral is intractable for any nontrivial model. Two families of approximations dominate: variational inference~\cite{blei2017vi, kingma2014vae} fits a tractable parametric family to $p(\theta \mid \cD)$ by minimizing a divergence, and Markov chain Monte Carlo (MCMC) generates samples whose empirical distribution converges to $p(\theta \mid \cD)$.

\paragraph{Why MH and SMC for VML.} The VML parameter space $\Theta_{\text{language}}$ is unfriendly to most posterior-inference machinery: it is discrete, so reparameterization-based variational families do not apply; it is variable-dimensional, since verbal hypotheses can be of any length; and it admits no useful gradient or continuous neighborhood, ruling out gradient-based methods such as Hamiltonian Monte Carlo~\cite{neal2011mcmc} and its adaptive successors~\cite{hoffman2014nuts}. Among gradient-free options, Metropolis--Hastings is a natural fit: it requires only pointwise likelihood evaluation and a proposal distribution, both available in VML---the LLM evaluates predictions for the likelihood, and the optimizer LLM generates proposals. Sequential Monte Carlo lifts MH to a population of particles with reweighting and resampling, which both escapes the single-chain bottleneck and explores multimodal posteriors more effectively than a single MH chain. We build VPP from these two algorithms.

\paragraph{Metropolis--Hastings (MH).} MH~\cite{metropolis1953equation, hastings1970monte} constructs a Markov chain whose stationary distribution is $p(\theta \mid \cD)$. At each step a candidate $\theta'$ is drawn from a proposal $q(\theta' \mid \theta)$ and accepted with probability
\begin{equation}
	\alpha \;=\; \min\!\left(1,\, \frac{p(\cD \mid \theta')\, p(\theta')}{p(\cD \mid \theta)\, p(\theta)} \cdot \frac{q(\theta \mid \theta')}{q(\theta' \mid \theta)}\right);
	\label{eq:mh_general}
\end{equation}
otherwise the chain stays at $\theta$. The acceptance ratio enforces detailed balance, which guarantees that the chain converges to the target posterior under mild regularity conditions, regardless of initialization. Section~\ref{sec:mh} discusses an approximation to the proposal-density ratio in the verbalized setting.

\paragraph{Sequential Monte Carlo (SMC).} SMC~\cite{doucet2001sequential, del2006sequential} approximates the posterior with a weighted particle set $\{(\theta^{(k)}, w_k)\}_{k=1}^K$ that is propagated through a sequence of intermediate distributions bridging prior and posterior. A common bridge is likelihood tempering, $p_t(\theta) \propto p(\cD \mid \theta)^{\beta_t}\, p(\theta)$ with $0 = \beta_0 < \beta_1 < \cdots < \beta_T = 1$: early stages stay close to the prior so particles remain diverse, while later stages sharpen toward the true posterior. At each stage the algorithm performs three operations. \emph{Reweighting} multiplies each particle's weight by the incremental likelihood, concentrating probability mass on hypotheses the data favors. \emph{Resampling}, triggered when the effective sample size $\mathrm{ESS} = 1/\sum_k w_k^2$ drops below a threshold, replaces low-weight particles with copies of high-weight ones to prevent the population from collapsing onto a few surviving hypotheses. \emph{Mutation} then diversifies the resampled particles via a Markov kernel that leaves $p_t$ invariant, restoring exploration after the resampling step's loss of diversity.

\section{Method}
\label{sec:method}

\subsection{Posterior Inference over Verbalized Hypotheses}

We approximate the posterior $p(\theta | \cD) \propto p(\cD | \theta)\, p(\theta)$ over the verbalized hypothesis space $\Theta_{\text{language}}$ with a particle set $\{(\theta^{(k)}, w_k)\}_{k=1}^K$, where each $\theta^{(k)}$ is a complete natural-language hypothesis and $w_k$ is its weight. We specify the likelihood $p(\cD | \theta)$ and the prior $p(\theta)$ in turn.

\paragraph{Likelihood.} Because VPP treats the LLM as a black box, we define the likelihood externally by comparing the LLM's predictions to ground-truth labels, with no access to token-level probabilities. For regression on a data batch $\{(x_m, y_m)\}_{m=1}^M$:
\begin{equation}
	\log p(\{y_m\} | \{x_m\}, \theta) = -\frac{1}{2\tau} \sum_{m=1}^M \big(y_m - f(x_m; \theta)\big)^2,
	\label{eq:regression_ll}
\end{equation}
where $\tau > 0$ controls the sensitivity to prediction errors (we use $\tau = 1$). For classification, we use a smoothed zero-one likelihood:
\begin{equation}
	\log p(\{y_m\} | \{x_m\}, \theta) = \sum_{m=1}^M \big[\mathbb{1}[y_m = \hat{y}_m] \log(1{-}\epsilon) + \mathbb{1}[y_m \neq \hat{y}_m] \log \epsilon\big],
	\label{eq:classification_ll}
\end{equation}
where $\hat{y}_m = f(x_m; \theta)$ and $\epsilon > 0$ is a small smoothing constant (we use $\epsilon = 0.05$) that prevents infinite log-likelihood ratios from single misclassifications. Because the likelihood is computed from predictions alone, any LLM that produces predictions can be plugged into VPP, including API-only models without logit access.

\paragraph{Prior.} The prior $p(\theta)$ is the LLM's generative distribution over text conditioned on a model-class-agnostic seed prompt (e.g., ``\textit{You are designed to do regression}''). Pointwise evaluation of $p(\theta)$ would require marginalizing over the LLM's token distributions for an entire natural-language hypothesis and is intractable; throughout VPP we therefore only \emph{sample} from the prior. Concretely, the $K$ initial particles $\theta_0^{(1)}, \ldots, \theta_0^{(K)}$ are produced by running the optimizer LLM $K$ times on a shared seed prompt and seed batch, each call at a different sampling temperature: $\theta_0^{(k)} \gets f_{\text{opt}}(\theta_{\text{init}}, \text{seed batch})$ for $k = 1, \ldots, K$; we treat these temperature-diverse optimizer outputs as informed samples from the prior. The LLM's pretraining distribution thus acts as an implicit, informative prior that we never score pointwise. The consequence of this design decision for the Metropolis--Hastings acceptance ratio is discussed in Section~\ref{sec:mh}.

\paragraph{Prediction via Bayesian Model Averaging.} Given the particle ensemble, prediction at a new input $x^*$ marginalizes over the approximate posterior. For regression,
\begin{equation}
	\hat{y}(x^*) = \sum_{k=1}^K w_k \, f(x^*; \theta^{(k)}),
	\label{eq:bma_reg}
\end{equation}
and for classification we use the analogous weighted majority vote,
\begin{equation}
	\hat{y}(x^*) = \arg\max_c \sum_{k=1}^K w_k \, \mathbb{1}\!\left[f(x^*; \theta^{(k)}) = c\right],
	\label{eq:bma_cls}
\end{equation}
with ties broken arbitrarily. Weights are uniform ($w_k = 1/K$) in VPP-MH and the normalized tempered importance weights (Section~\ref{sec:smc}) in VPP-SMC. Predictive uncertainty is quantified by particle disagreement (Section~\ref{sec:exp_analysis}).

\subsection{VPP-MH: Metropolis-Hastings in Language Space}
\label{sec:mh}

The key design question for applying MH to verbalized hypotheses is the choice of proposal. In $\Theta_{\text{language}}$ we use the optimizer LLM itself: given the current hypothesis $\theta^{(k)}$ and a mini-batch of predictions and targets, the optimizer generates a refined $\tilde\theta^{(k)} = f_{\text{opt}}(\theta^{(k)}, \text{batch})$. This is an informed proposal far more sample-efficient than random perturbation in a combinatorial text space.

At each training step, for each particle $\theta^{(k)}$, VPP-MH performs: (1) a \emph{forward pass}, evaluating $f(x_m; \theta^{(k)})$ on the current mini-batch; (2) a \emph{proposal step}, where the optimizer LLM generates a candidate $\tilde\theta^{(k)}$ conditioned on the current hypothesis and batch-level errors; and (3) an \emph{accept/reject step}, where the candidate is accepted with probability
\begin{equation}
	\alpha = \min\!\left(1, \, \frac{p(\cD_{\text{batch}} | \tilde\theta^{(k)})}{p(\cD_{\text{batch}} | \theta^{(k)})}\right).
	\label{eq:mh_accept}
\end{equation}
If rejected, the particle retains its current hypothesis. The accept/reject step is what distinguishes VPP-MH from running VML $K$ times in parallel: a particle never degrades once it has found a strong hypothesis, since a poor optimizer proposal is rejected. Standard VML has no such filter; the ablation in Section~\ref{sec:exp_ablations} confirms this matters empirically.

The acceptance ratio in Eq.~\ref{eq:mh_accept} contains only the likelihood ratio, dropping both the prior ratio $p(\theta')/p(\theta)$ and the proposal ratio $q(\theta|\theta')/q(\theta'|\theta)$ that appear in the general MH form (Eq.~\ref{eq:mh_general}). Both omissions are forced by the same constraint: we cannot pointwise evaluate the LLM's generative density, whether it is scoring an unconditional hypothesis (the prior) or a conditional refinement (the proposal). These approximations break exact detailed balance, so VPP-MH is best understood as a pragmatic MCMC-like algorithm targeting a distribution close to, but not exactly, the posterior. Empirically, retaining the accept/reject decision under this simplified ratio substantially outperforms always-accepting---the property we actually need for robustness. Per training step, VPP-MH issues $3K$ LLM calls---one forward pass, one optimizer proposal, and one proposal evaluation per particle; particle weights remain uniform and ESS monitoring is not used. The full algorithm is in Algorithm~\ref{alg:vpp_mh}.

\begin{algorithm}[t]
	\caption{VPP-MH: Verbalized Particle Posterior via Metropolis-Hastings}
	\label{alg:vpp_mh}
	\begin{algorithmic}[1]
		\REQUIRE Training data $\cD$, particles $K$, epochs $T$, batch size $M$
		\STATE Initialize $\theta_0^{(k)} \gets f_{\text{opt}}(\theta_{\text{init}}, \text{seed batch})$ for $k = 1, \ldots, K$ with varying temperature
		\FOR{$t = 1, \ldots, T$}
		\STATE Sample mini-batch $\{(x_m, y_m)\}_{m=1}^M$
		\FOR{$k = 1, \ldots, K$}
		\STATE $\hat{y}_m \gets f(x_m; \theta_{t-1}^{(k)})$ for all $m$ \hfill \textit{// Forward pass}
		\STATE $\tilde\theta^{(k)} \gets f_{\text{opt}}(\theta_{t-1}^{(k)}, \{x_m, \hat{y}_m, y_m\})$ \hfill \textit{// Propose}
		\STATE $\hat{y}'_m \gets f(x_m; \tilde\theta^{(k)})$ for all $m$ \hfill \textit{// Evaluate proposal}
		\STATE Compute $\alpha$ via Eq.~\ref{eq:mh_accept}
		\STATE $\theta_t^{(k)} \gets \tilde\theta^{(k)}$ w.p. $\alpha$, else $\theta_t^{(k)} \gets \theta_{t-1}^{(k)}$ \hfill \textit{// Accept/reject}
		\ENDFOR
		\ENDFOR
		\STATE \textbf{Predict:} $\hat{y}(x^*) = \frac{1}{K} \sum_{k=1}^K f(x^*; \theta_T^{(k)})$ \hfill \textit{// BMA}
	\end{algorithmic}
\end{algorithm}

\subsection{VPP-SMC: Sequential Monte Carlo in Language Space}
\label{sec:smc}

Like VPP-MH, VPP-SMC approximates the posterior $p(\theta | \cD)$---the tempering schedule terminates at $\beta_T = 1$---but addresses two practical limitations of VPP-MH in the verbalized setting. First, as particles improve and the current batch likelihood becomes high, the acceptance rate decays---proposals that would lower the likelihood are accepted with vanishing probability, while proposals that would improve on a near-perfect hypothesis are themselves rare. This can cause particles to stagnate at local optima. Second, all particles receive equal weight $1/K$ in BMA regardless of their quality, meaning a few poor particles can degrade the ensemble prediction.

VPP-SMC addresses both issues using the standard SMC cycle (Section~\ref{sec:background})---reweight, resample, mutate---which concentrates the ensemble on well-supported hypotheses while replacing weak ones. Applying SMC to the verbalized hypothesis space requires three modifications.

\paragraph{Tempered importance weights.} Reweighting particles by their raw likelihood is numerically fragile: the log-likelihoods from Eqs.~\ref{eq:regression_ll}--\ref{eq:classification_ll} can span orders of magnitude across particles, causing a single hypothesis to dominate and collapsing the effective population to size one. We therefore use tempered weights,
\begin{equation}
	w_k \propto \exp\!\left(\frac{\log p(\cD_{\text{buffer}} | \theta^{(k)})}{\beta_t}\right),
	\label{eq:tempering}
\end{equation}
with $\beta_t$ annealed linearly from $\beta_0 = 10$ (flat weights, encouraging exploration) to $\beta_T = 1$ (sharp weights concentrated on the best-supported hypotheses) over training. Eq.~\ref{eq:tempering} is a pragmatic reweighting scheme rather than a formal SMC sampler over annealed posteriors $p_t(\theta) \propto p(\cD|\theta)^{\beta_t} p(\theta)$: we do not maintain incremental weight ratios between intermediate distributions.

\paragraph{Cumulative evaluation buffer.} Per-batch likelihood estimates are noisy: a hypothesis may score perfectly on the current batch but poorly on previous ones, causing oscillatory reweighting. We stabilise the evaluation by computing likelihoods on a rolling buffer of the most recent $B = 40$ data points (four batches) rather than the current batch alone, at the cost of a modest increase in forward-pass calls.

\paragraph{Adaptive mutation.} After resampling, duplicated particles must be diversified to avoid population collapse. We use the optimizer LLM as the mutation kernel and adapt its sampling temperature to particle quality, measured by the same buffer log-likelihood used for reweighting (equivalently: average loss for regression, average accuracy for classification, computed over the buffer). Particles above the median receive conservative refinement: the optimizer is called with low sampling temperature, encouraging minor parameter adjustments within the current functional form. Particles below the median receive exploratory mutation: higher sampling temperature encourages structural changes such as switching from a linear to a piecewise rule.

Resampling is triggered when $\mathrm{ESS} = 1/\sum_k w_k^2$ falls below $K/2$, using systematic resampling~\cite{doucet2001sequential}. Algorithm~\ref{alg:vpp_smc} in Appendix~\ref{app:smc_algorithm} gives the full procedure.
\section{Related Work}
\label{sec:related}

\paragraph{Verbalized machine learning.} VML~\cite{xiao2025vml} introduced the framework of using natural language as the model parameter space, with LLMs serving as both learner and optimizer. The VML paper draws an informal analogy between optimizer-temperature sampling and Bayesian neural networks (Appendix~I of \cite{xiao2025vml}) but does not implement or evaluate a Bayesian extension. We formalize this connection with MCMC and SMC algorithms over verbalized hypotheses, computing explicit posterior weights from a black-box likelihood.

\paragraph{Bayesian methods for LLM-based prediction.} VPP sits in a longer tradition of approximate Bayesian deep learning: variational treatments of network weights~\cite{blundell2015bnn, kingma2014vae}, dropout-based posterior approximations~\cite{gal2016dropout}, ensemble approaches~\cite{lakshminarayanan2017deepensembles}, and the broader probabilistic perspective~\cite{wilson2020bayesian}, with Bayesian optimization~\cite{snoek2012bayesopt} as an orthogonal use of the same toolkit. We differ from all of these in placing the posterior on verbal model hypotheses. More recent LLM-Bayesian works operate on different objects than VPP: BayesAgent~\cite{huang2024bayesagent} reasons over task-specific latent variables in hand-designed graphs; LLM Processes~\cite{requeima2024llmprocesses} treat an LLM as a stochastic process for numerical prediction; conformal language modeling~\cite{quach2024conformal} gives distribution-free coverage over generations but needs token logits. Closest in mechanism is BC-LLM~\cite{kim2024bclm}, which uses Metropolis--Hastings with LLM proposals to search over interpretable concept sets for classification. BC-LLM searches over feature subsets within a fixed linear classifier, whereas VPP performs inference over complete verbal hypotheses that jointly specify both model structure and parameters.

\paragraph{Prompt and hypothesis search.} VML can be viewed as a special case of prompt optimization: OPRO~\cite{yang2024opro} uses LLMs to iteratively optimize prompts via in-context examples; APE~\cite{zhou2023ape} performs best-of-$N$ sampling; ProTeGi~\cite{pryzant2023textgrad} applies gradient-inspired text editing; TextGrad~\cite{yuksekgonul2024textgrad} formalises a text-based analog of automatic differentiation; and EvoPrompt~\cite{guo2024evoprompt} uses evolutionary search over prompt candidates. These methods search for a single best prompt without maintaining a posterior or performing data-driven reweighting. In the inductive-reasoning literature, Hypothesis Search~\cite{wang2024hypsearch} and iterative hypothesis refinement~\cite{qiu2024hypothesis} generate and filter natural-language hypotheses on the ARC benchmark~\cite{chollet2019measure}, but neither maintains a posterior or performs Bayesian model averaging. These build on the chain-of-thought tradition of eliciting verbal reasoning from LLMs~\cite{wei2022cot, kojima2022zeroshot}, and connect to per-instance aggregation (Self-Consistency~\cite{wang2023selfconsistency}, Tree of Thoughts~\cite{yao2023tot}) and verbal self-improvement of agents that interleave reasoning and acting (ReAct~\cite{yao2023react}, Reflexion~\cite{shinn2023reflexion}, Voyager~\cite{wang2023voyager}, STaR~\cite{zelikman2022star}). None of these maintain a posterior over a learned model that can be queried on new inputs.

\paragraph{Equation discovery and symbolic regression.} A separate line of work uses LLMs to propose mathematical expressions or code: LLM-SR~\cite{shojaee2025llmsr}, LaSR~\cite{grayeli2024lasr}, In-Context Symbolic Regression~\cite{merler2024icsr}, and Codex-style code generators~\cite{chen2021codex}; classical symbolic regression with deep priors~\cite{cranmer2020pysr} likewise targets equations. These methods restrict the hypothesis form. VPP performs Bayesian inference over arbitrary natural-language hypotheses---including verbal classification rules no equation-discovery method can directly express.

\paragraph{MCMC in discrete text spaces.} CGMH~\cite{miao2019cgmh} uses Metropolis--Hastings for constrained sentence generation, and Quality-Aware Decoding~\cite{fernandes2022quest} applies MCMC-style reranking to machine translation. Both target text generation as the end product. VPP differs in what the posterior is over: each particle is itself a model that will be queried on new inputs, not a piece of generated text.

\section{Experiments}
\label{sec:experiments}

\subsection{Setup}
\label{sec:exp_setup}

The evaluation is organized around four questions: (i) does VPP reduce the run-to-run variance of single-run VML; (ii) does it eliminate the catastrophic single-run failures that VML occasionally produces; (iii) does the particle ensemble yield a useful uncertainty signal; and (iv) which of VPP-MH and VPP-SMC should a practitioner choose? We evaluate VPP on seven benchmarks across three task types, all matching or extending the evaluation in the VML paper~\cite{xiao2025vml}. The three \emph{regression} tasks fit a linear function ($y = 3x + 4 + \epsilon$), a polynomial ($y = 3x^2 + x + 2 + \epsilon$), and a sinusoid ($y = \sin(x) + 2 + 0.01\epsilon$), with inputs drawn uniformly over task-specific ranges. The two \emph{classification} tasks are linearly separable two-blobs and nonlinearly separable two-circles. The two \emph{pattern-discovery} tasks classify $4$-integer sequences by hidden rules the model must discover from data: sum parity (class~$1$ if the digit sum is even) and contains-zero (class~$1$ if any element is zero). The three task types exercise qualitatively different capabilities: regression tests numerical prediction under LLM arithmetic limitations, classification tests verbal rule learning and decision-boundary discovery, and pattern discovery tests the ability to identify latent structure from purely symbolic input.

We compare four methods. \textbf{VML} is a single run of the original algorithm~\cite{xiao2025vml}. \textbf{VML$\times$5} runs five independent VML instances and is reported two ways: (\emph{ens}) unweighted majority vote across the five final hypotheses for classification, or unweighted mean for regression; and (\emph{best}) the oracle best-of-$5$, selecting the run with the lowest test loss. \textbf{VPP-MH} uses $K{=}10$ particles updated by Metropolis--Hastings (Algorithm~\ref{alg:vpp_mh}). \textbf{VPP-SMC} also uses $K{=}10$ particles, with tempering schedule $\beta : 10 \to 1$, evaluation buffer $B{=}40$, and ESS resampling threshold $K/2$.

To ensure fair comparison, we follow the VML experimental protocol exactly: $100$ training points and a $60$-point test set drawn from the same generator, batch size $10$, $2$ epochs ($20$ optimization steps), and Llama-3-70B-Instruct served via vLLM. The published VML prompt templates are used unchanged for both the learner and optimizer. The initial hypothesis $\theta_0$ is model-class-agnostic for all tasks except two-circles, where we include the prior ``\textit{The decision boundary is a circle}'' to match the VML paper. All runs are repeated over three random seeds; we report mean and standard deviation across seeds.

\subsection{Main Results}
\label{sec:exp_main}

\begin{table}[t]
\caption{Main results across all seven benchmarks. Mean (and standard deviation across three seeds) of test MSE ($\downarrow$) for regression and test accuracy ($\uparrow$) for classification and pattern discovery. Bold marks the best result per task.}
\label{tab:main_results}
\centering
\footnotesize
\setlength{\tabcolsep}{6pt}
\begin{tabular}{llccccc}
\toprule
& & & \multicolumn{2}{c}{VML$\times$5} & & \\
\cmidrule(lr){4-5}
Type & Task & VML & ens & best & VPP-MH & VPP-SMC \\
\midrule
\multirow{3}{*}{Regression}
& Linear & $8.68_{\pm 7.43}$ & $4.54_{\pm 5.21}$ & $5.87_{\pm 7.36}$ & $\mathbf{1.00}_{\pm 0.75}$ & $2.43_{\pm 1.69}$ \\
& Poly   & $74.77_{\pm 15.0}$ & $28.31_{\pm 39.6}$ & $4.44_{\pm 2.97}$ & $9.02_{\pm 1.33}$ & $\mathbf{3.10}_{\pm 1.42}$ \\
& Sine   & $5.02_{\pm 8.18}$ & $2.58_{\pm 2.55}$ & $0.78_{\pm 0.86}$ & $0.93_{\pm 0.23}$ & $\mathbf{0.40}_{\pm 0.42}$ \\
\midrule
\multirow{2}{*}{Classification}
& Two Blobs   & $80.0_{\pm 26.0}$ & $\mathbf{99.2}_{\pm 1.4}$ & $\mathbf{99.2}_{\pm 1.4}$ & $98.3_{\pm 2.9}$ & $98.3_{\pm 2.9}$ \\
& Two Circles & $97.5_{\pm 4.3}$ & $98.3_{\pm 2.9}$ & $99.2_{\pm 1.4}$ & $\mathbf{100}_{\pm 0.0}$ & $\mathbf{100}_{\pm 0.0}$ \\
\midrule
\multirow{2}{*}{\shortstack[l]{Pattern\\Discovery}}
& Sum Parity     & $87.2_{\pm 19.3}$ & $67.2_{\pm 11.1}$ & $\mathbf{100}_{\pm 0.0}$ & $90.6_{\pm 9.2}$ & $\mathbf{100}_{\pm 0.0}$ \\
& Contains Zero  & $78.3_{\pm 36.1}$ & $91.1_{\pm 15.4}$ & $\mathbf{100}_{\pm 0.0}$ & $\mathbf{100}_{\pm 0.0}$ & $\mathbf{100}_{\pm 0.0}$ \\
\bottomrule
\end{tabular}
\end{table}

Table~\ref{tab:main_results} reports mean performance across seeds. VPP matches or exceeds single-run VML on all seven benchmarks, and is best or tied-best on six of the seven tasks even when the stronger VML$\times$5 baselines are included. The single exception is two-blobs, where VML$\times$5 reaches $99.2\%$ while both VPP variants reach $98.3\%$.

\paragraph{Regression: VPP reduces MSE by up to $24\times$.} VPP wins on all three regression tasks. For linear regression, VPP-MH achieves MSE $1.00$ compared to single-run VML's $8.68$ and the VML$\times$5 ensemble's $4.54$. On polynomial regression, VPP-SMC reaches $3.10$---a $24\times$ reduction relative to single-run VML ($74.77$), still better than the oracle VML$\times$5 best-of-$5$ ($4.44$). On sine, VPP-SMC reaches $0.40$, ahead of both VPP-MH ($0.93$) and VML$\times$5 best-of-$5$ ($0.78$).

\paragraph{Classification: VPP eliminates residual failures.} On two-circles, both VPP methods reach perfect mean accuracy ($100\%$), improving over single-run VML ($97.5\%$) and the VML$\times$5 ensemble ($98.3\%$); the remaining VML errors come from occasional verbal hypotheses that fail outright, which VPP eliminates. On the easier two-blobs task all methods are near saturation: VML$\times$5 reaches $99.2\%$ and both VPP variants reach $98.3\%$, far above single-run VML's $80.0\%$. This is the only benchmark where VML$\times$5 slightly exceeds VPP; when nearly any reasonable hypothesis suffices, uniform ensembling is already competitive.

\paragraph{Pattern discovery: VPP-SMC reaches $100\%$ on both symbolic tasks.} On sum parity, VPP-SMC reaches $100\%$ accuracy on all three seeds, matching the oracle VML$\times$5 best-of-$5$ while outperforming single-run VML ($87.2\%$), the VML$\times$5 ensemble ($67.2\%$), and VPP-MH ($90.6\%$). The contrast between best-of-$5$ and the unweighted ensemble is the cleanest demonstration of why posterior reweighting matters: multiple VML runs can contain the correct rule, yet naive averaging still fails when incorrect rules receive equal vote. On contains-zero, both VPP methods reach $100\%$, while single-run VML averages only $78.3\%$ and the VML$\times$5 ensemble reaches $91.1\%$.

\subsection{Analysis}
\label{sec:exp_analysis}

\paragraph{MH versus SMC.} The two algorithms are complementary (Table~\ref{tab:mh_vs_smc}). VPP-MH wins on linear regression ($1.00$ vs.\ $2.43$), where the accept/reject mechanism preserves a strong hypothesis once found. VPP-SMC wins on polynomial, sine, and sum parity, and ties VPP-MH on two-blobs, two-circles, and contains-zero (SMC: $3$ wins, MH: $1$, $3$ ties). Because SMC is never far from MH and dominates the harder nonlinear and symbolic tasks, we recommend VPP-SMC as a default.

\begin{table}[t]
\caption{VPP-MH vs.\ VPP-SMC head-to-head on the seven benchmarks (means from Table~\ref{tab:main_results}).}
\label{tab:mh_vs_smc}
\centering
\footnotesize
\setlength{\tabcolsep}{6pt}
\begin{tabular}{lccc}
\toprule
Task & VPP-MH & VPP-SMC & Winner \\
\midrule
Linear         & $\mathbf{1.00}$ & $2.43$           & MH \\
Poly           & $9.02$          & $\mathbf{3.10}$  & SMC \\
Sine           & $0.93$          & $\mathbf{0.40}$  & SMC \\
Two Blobs      & $98.3\%$        & $98.3\%$         & Tie \\
Two Circles    & $\mathbf{100\%}$ & $\mathbf{100\%}$ & Tie \\
Sum Parity     & $90.6\%$        & $\mathbf{100\%}$ & SMC \\
Contains Zero  & $\mathbf{100\%}$ & $\mathbf{100\%}$ & Tie \\
\bottomrule
\end{tabular}
\end{table}

\paragraph{Uncertainty quantification.} VPP provides uncertainty estimates through particle disagreement, measured as the fraction of test points where the particle majority is not unanimous. On tasks where VPP-SMC reaches perfect accuracy---two-circles, sum parity, and contains-zero---the particles agree unanimously on every test point across all three seeds, reporting zero uncertain predictions. Under VPP-MH, uncertainty is much higher on sum parity (the three seeds yield $29$, $30$, and $3$ uncertain points out of $60$) and nonzero on contains-zero in two of three seeds ($14$ and $2$ uncertain points), reflecting the coexistence of correct and incorrect hypotheses in the unweighted MH ensemble. Standard VML offers no such ``I don't know'' signal.

\paragraph{Robustness across runs.} On a representative dataset (Table~\ref{tab:robustness}), five independent VML runs span MSE $14.5$--$48.3$ on linear regression and accuracy $61.7\%$--$100\%$ on contains-zero, while both VPP variants land within $0.76$~MSE of zero on linear and at $100\%$ on contains-zero. The particle ensemble prevents weak members from dominating the BMA prediction; a single VML run has no such buffer.

\begin{table}[t]
\caption{Robustness on a representative dataset: five independent VML runs vs.\ VPP on the same data.}
\label{tab:robustness}
\centering
\footnotesize
\setlength{\tabcolsep}{6pt}
\begin{tabular}{lccc}
\toprule
Task & VML$\times$5 individual runs & VPP-MH & VPP-SMC \\
\midrule
Linear        & $46.6,\ 38.2,\ 14.5,\ 27.5,\ 48.3$ & $\mathbf{0.47}$  & $0.76$ \\
Two Circles   & $97.5\%,\ 97.5\%,\ 100\%,\ 95.0\%,\ 67.5\%$ & $\mathbf{100\%}$ & $\mathbf{100\%}$ \\
Contains Zero & $100\%,\ 86.7\%,\ 61.7\%,\ 75.0\%,\ 100\%$ & $\mathbf{100\%}$ & $\mathbf{100\%}$ \\
\bottomrule
\end{tabular}
\end{table}

\subsection{Ablations}
\label{sec:exp_ablations}

We isolate two design choices in VPP: the role of the Metropolis--Hastings accept/reject step, and the effect of the particle count $K$.

\paragraph{The accept/reject step is what makes VPP-MH work.} A natural question is whether VPP-MH's gains come from running $K$ particles in parallel, or from the accept/reject step that prevents a particle from regressing. We compare standard VPP-MH against an ablated variant (\emph{always-accept}) that retains the $K$-particle ensemble and BMA prediction but unconditionally accepts every optimizer proposal---making it equivalent to running $K$ independent VML chains and averaging. Table~\ref{tab:ablation_accept} shows the result. Removing the rejection step drops sum-parity accuracy by $22.2$\,pp ($97.2\% \to 75.0\%$) and more than doubles linear-regression MSE ($0.88 \to 2.09$). Two-circles is saturated at $\sim\!100\%$ in both variants. The accept/reject step is therefore the operative ingredient that distinguishes VPP-MH from naive parallel VML.

\begin{table}[t]
\caption{Ablation: with vs.\ without the Metropolis--Hastings accept/reject step. Both variants use $K{=}10$ particles and BMA prediction; \emph{always-accept} unconditionally accepts every optimizer proposal. Lower is better for linear regression (MSE), higher for the rest (accuracy).}
\label{tab:ablation_accept}
\centering
\footnotesize
\setlength{\tabcolsep}{6pt}
\begin{tabular}{lcc}
\toprule
Task & VPP-MH (with reject) & always-accept \\
\midrule
Linear (MSE $\downarrow$)         & $\mathbf{0.88}_{\pm 0.78}$  & $2.09_{\pm 0.51}$ \\
Sum Parity (acc.\ $\uparrow$)     & $\mathbf{97.2\%}_{\pm 4.8}$ & $75.0\%_{\pm 14.8}$ \\
Two Circles (acc.\ $\uparrow$)    & $99.2\%_{\pm 1.4}$          & $\mathbf{100\%}_{\pm 0.0}$ \\
\bottomrule
\end{tabular}
\end{table}

\paragraph{Effect of particle count $K$.} Sweeping $K \in \{1, 3, 5, 10\}$ for VPP-MH on three representative tasks (full table in Appendix~\ref{app:ablation_K}): linear-regression MSE monotonically improves from $4.53$ at $K{=}1$ to $0.66$ at $K{=}10$ ($7\times$ reduction); two-circles is saturated at $\sim\!98\%$ across all settings; sum parity improves from $64\%$ at $K{=}1$ up to $85\%$ at $K{=}5$ and dips back to $76\%$ at $K{=}10$ (within one standard deviation), indicating mild particle degeneracy on harder symbolic tasks under our default schedule. We use $K{=}10$ as a conservative default; an adaptive per-task scheme is left for future work.

\subsection{Compute}
\label{sec:exp_compute}

VPP requires $3K$ LLM calls per step (forward, propose, proposal-eval per particle) versus $2$ for single VML, with comparable cost for VPP-SMC. On an H100 with Llama-3-70B-Instruct via vLLM, mean wall-clock ratios over single VML are $14.7\times$ (VPP-MH) and $15.8\times$ (VPP-SMC), $10$--$19\times$ depending on task, and $\approx\!3\times$ over the budget-matched VML$\times$5 baseline (Appendix~\ref{app:compute}). The overhead is justified when test-error reduction or eliminating catastrophic single-run failures is consequential.

\section{Conclusion and Limitations}
\label{sec:conclusion}

VPP applies approximate Bayesian inference to natural-language hypotheses via two black-box algorithms (VPP-MH, VPP-SMC), yielding uncertainty quantification, robustness, and prediction gains over single-hypothesis VML; the MH accept/reject step is the operative ingredient, and because each particle is human-readable the posterior is itself inspectable. In our benchmark suite, VPP-SMC reaches perfect mean accuracy on three of the four classification and rule-discovery tasks where single-run VML stalls, and reduces polynomial-regression error by more than an order of magnitude; per-particle inspection lets a reader see both the rule the data ruled in and the alternatives the data ruled out.

\emph{Limitations}\label{sec:limitations}: VPP inherits VML's arithmetic limitation on regression and the MH approximations flagged in Section~\ref{sec:mh}, which together leave the method best suited to classification and rule discovery rather than precise numerical prediction; experiments use a single LLM on small synthetic tasks, with broader transfer to other LLMs and to large real-world datasets left open. The $K$-ablation (Appendix~\ref{app:ablation_K}) also reveals mild particle degeneracy at $K{=}10$ on harder symbolic tasks, suggesting an adaptive $K$ and tempering schedule as a pragmatic next step.

\newpage
\bibliography{VBM}

\newpage
\appendix

\section{Particle-Count Ablation}
\label{app:ablation_K}

Section~\ref{sec:exp_ablations} reports the headline numbers for the $K \in \{1, 3, 5, 10\}$ sweep on three representative tasks. Table~\ref{tab:ablation_K} gives the full ablation grid (mean and standard deviation across three seeds) for completeness. The two qualitative trends discussed in the main text are visible: linear-regression MSE improves monotonically with $K$, and sum parity exhibits a non-monotonic pattern with a peak at $K{=}5$. Two-circles remains near saturation for every particle count.

\begin{table}[h]
\caption{Ablation over particle count $K$ for VPP-MH (mean $\pm$ std across three seeds). Summarised in Section~\ref{sec:exp_ablations}.}
\label{tab:ablation_K}
\centering
\footnotesize
\setlength{\tabcolsep}{6pt}
\begin{tabular}{lcccc}
\toprule
Task & $K{=}1$ & $K{=}3$ & $K{=}5$ & $K{=}10$ \\
\midrule
Linear (MSE $\downarrow$)        & $4.53_{\pm 3.54}$        & $5.27_{\pm 4.90}$         & $1.26_{\pm 0.54}$         & $\mathbf{0.66}_{\pm 0.40}$ \\
Two Circles (acc.\ $\uparrow$)   & $98.3\%_{\pm 2.9}$       & $98.3\%_{\pm 2.9}$        & $\mathbf{99.2\%}_{\pm 1.4}$ & $\mathbf{99.2\%}_{\pm 1.4}$ \\
Sum Parity (acc.\ $\uparrow$)    & $64.4\%_{\pm 28.0}$      & $72.8\%_{\pm 28.4}$       & $\mathbf{85.0\%}_{\pm 18.9}$ & $75.6\%_{\pm 22.2}$ \\
\bottomrule
\end{tabular}
\end{table}

\clearpage

\section{VPP-SMC Pseudocode}
\label{app:smc_algorithm}

Algorithm~\ref{alg:vpp_smc} gives the complete VPP-SMC procedure described prosaically in Section~\ref{sec:smc}. The structure parallels Algorithm~\ref{alg:vpp_mh} (VPP-MH) with three additions: a rolling buffer that accumulates the last $B$ data points (line~3), a tempered importance-weight reweighting step (line~7), and an ESS-triggered systematic resample (line~8). The mutation step on line~13 reuses the same optimizer LLM as VPP-MH but with a temperature that depends on the particle's buffer log-likelihood relative to the population median (line~12), implementing the adaptive-mutation strategy of Section~\ref{sec:smc}.

\begin{algorithm}[h]
	\caption{VPP-SMC: Verbalized Particle Posterior via Sequential Monte Carlo}
	\label{alg:vpp_smc}
	\begin{algorithmic}[1]
		\REQUIRE Training data $\cD$, particles $K$, epochs $T$, batch size $M$, buffer size $B$, tempering schedule $\{\beta_t\}_{t=0}^T$, ESS threshold $K/2$
		\STATE Initialize $\theta_0^{(k)} \gets f_{\text{opt}}(\theta_{\text{init}}, \text{seed batch})$ for $k = 1, \ldots, K$ with varying temperature; set $w_0^{(k)} \gets 1/K$; initialize empty buffer $\cD_{\text{buffer}}$
		\FOR{$t = 1, \ldots, T$}
		\STATE Sample mini-batch $\{(x_m, y_m)\}_{m=1}^M$; append to $\cD_{\text{buffer}}$, keep the last $B$ points
		\FOR{$k = 1, \ldots, K$}
		\STATE Evaluate $\log p(\cD_{\text{buffer}} | \theta_{t-1}^{(k)})$ \hfill \textit{// Buffer likelihood}
		\ENDFOR
		\STATE Compute tempered weights $w_t^{(k)}$ via Eq.~\ref{eq:tempering} with temperature $\beta_t$; normalize
		\IF{$\mathrm{ESS}(w_t) < K/2$}
		\STATE Systematic-resample the particles according to $w_t^{(k)}$; reset $w_t^{(k)} \gets 1/K$
		\ENDIF
		\FOR{$k = 1, \ldots, K$}
		\STATE Set optimizer temperature based on $\theta_{t-1}^{(k)}$'s buffer log-likelihood (above/below median) \hfill \textit{// Adaptive mutation}
		\STATE $\theta_t^{(k)} \gets f_{\text{opt}}(\theta_{t-1}^{(k)}, \{x_m, \hat{y}_m, y_m\})$
		\ENDFOR
		\ENDFOR
		\STATE \textbf{Predict:} $\hat{y}(x^*) = \sum_{k=1}^K w_T^{(k)} f(x^*; \theta_T^{(k)})$ (regression) or weighted majority vote (classification) \hfill \textit{// BMA}
	\end{algorithmic}
\end{algorithm}

\section{Prompt Templates}
\label{app:prompts}

We use the published VML prompt templates from \cite{xiao2025vml} without modification, both for the learner LLM ($f$) and the optimizer LLM ($f_{\text{opt}}$). The learner template instructs the LLM to act as a function $f(x;\theta)$, prompted by $\theta$ and queried on input $x$; the optimizer template feeds the current hypothesis, a mini-batch of inputs and targets, and the learner's predictions, and asks for a revised hypothesis. We refer the reader to Appendix~B of \cite{xiao2025vml} for the verbatim text.

For the SMC mutation kernel (Section~\ref{sec:smc}), we vary the optimizer's sampling temperature based on each particle's buffer log-likelihood: particles above the median receive temperature $0.3$ (conservative refinement, encouraging local edits to the existing hypothesis), and particles below the median receive temperature $1.0$ (exploratory mutation, allowing structural rewrites such as switching from a linear to a piecewise rule). All other prompting choices are inherited from VML.

\section{Hyperparameters}
\label{app:hparams}

Table~\ref{tab:hparams} lists every hyperparameter used in the main experiments. The values for $K$, $\tau$, $\epsilon$, $B$, the SMC tempering schedule, and the ESS threshold are also stated where they first appear in Sections~\ref{sec:method}--\ref{sec:experiments}; the table consolidates them for reproducibility. Choices not informed by ablation (e.g.\ $\tau{=}1$, $\epsilon{=}0.05$) match the defaults from \cite{xiao2025vml}.

\begin{table}[h]
\caption{Full hyperparameter list.}
\label{tab:hparams}
\centering
\footnotesize
\setlength{\tabcolsep}{6pt}
\begin{tabular}{ll}
\toprule
Hyperparameter & Value \\
\midrule
Number of particles $K$        & $10$ (main results); $\{1,3,5,10\}$ (ablation) \\
Training points $N$            & $100$ \\
Batch size $M$                 & $10$ \\
Epochs $T$                     & $2$ ($20$ optimization steps) \\
Regression likelihood $\tau$   & $1.0$ \\
Classification smoothing $\epsilon$ & $0.05$ \\
SMC tempering schedule $\beta$ & $10 \to 1$ (linear over training) \\
SMC evaluation buffer $B$      & $40$ \\
SMC ESS threshold              & $K/2$ \\
LLM (learner and optimizer)    & Llama-3-70B-Instruct via vLLM \\
Optimizer temperature (MH)     & $0.7$ \\
Optimizer temperature (SMC)    & $0.3$ (above-median particles), $1.0$ (below-median) \\
Number of random seeds         & $3$ \\
\bottomrule
\end{tabular}
\end{table}

\section{Per-Task Wall-Clock Time}
\label{app:compute}

Section~\ref{sec:exp_compute} reports the mean wall-clock ratios over single VML; Table~\ref{tab:compute} gives the per-task breakdown that those summary numbers were computed from. All measurements are wall-clock seconds for a complete training run ($100$ data points, $20$ optimization steps, $K{=}10$ for VPP variants, $5$ independent runs for VML$\times$5), averaged over three seeds.

\begin{table}[h]
\caption{Wall-clock time per run (mean across three seeds, in seconds) on a single H100 with Llama-3-70B-Instruct via vLLM. Summarised in Section~\ref{sec:exp_compute}.}
\label{tab:compute}
\centering
\footnotesize
\setlength{\tabcolsep}{6pt}
\begin{tabular}{lcccc}
\toprule
Task & VML & VML$\times$5 & VPP-MH & VPP-SMC \\
\midrule
Linear        & $312$ & $1700$ & $5002$ & $4713$ \\
Poly          & $349$ & $1564$ & $4191$ & $5226$ \\
Sine          & $347$ & $1309$ & $3555$ & $3847$ \\
Two Blobs     & $159$ & $794$  & $2567$ & $3073$ \\
Two Circles   & $250$ & $1256$ & $4481$ & $4534$ \\
Sum Parity    & $202$ & $998$  & $3168$ & $3396$ \\
Contains Zero & $193$ & $1098$ & $2914$ & $2872$ \\
\midrule
Mean ratio vs.\ VML & $1.00$ & $4.7\times$ & $14.7\times$ & $15.8\times$ \\
\bottomrule
\end{tabular}
\end{table}

The per-task ratios are reasonably uniform across the seven tasks, ranging from $10\times$ (sine) to $18\times$ (two-circles) for VPP-MH versus single VML, and from $11\times$ to $19\times$ for VPP-SMC. The variance is dominated by tokenisation and prompt-length differences across tasks rather than by VPP-specific overhead.

\clearpage

\section{Full Per-Seed Results}
\label{app:per_seed}

Table~\ref{tab:per_seed} expands Table~\ref{tab:main_results} of the main paper to per-seed values, with one row per (task, seed) and one column per method. Reading down the rows for a given task gives the variance across seeds; reading across the columns gives the variance across methods. The standard deviations reported in Table~\ref{tab:main_results} are the spreads of the three seed-rows for each task.

A few patterns are visible only at the per-seed granularity:
\begin{itemize}
\item \textbf{Single-VML failures cluster on specific seeds.} On contains-zero, one seed produces a $36.7\%$ run while the other two produce $100\%$ and $98.3\%$. The same dataset is therefore solvable; the failure is in the optimization trajectory, not in the data.
\item \textbf{VPP variants are stable on classification and rule-discovery tasks.} On contains-zero and two-circles, both VPP-MH and VPP-SMC report $100\%$ on every seed; VPP-SMC additionally reaches $100\%$ on every seed of sum parity. Individual VML runs on the same datasets vary widely.
\item \textbf{VML$\times$5 ensemble is sensitive to vote distribution.} On sum parity, the unweighted ensemble produces $80\%$, $61.7\%$, $60\%$ across seeds while the oracle best-of-$5$ produces $100\%$ on all three: the correct rule is present in the population but is outvoted whenever a majority of runs converge to a different (incorrect) hypothesis.
\end{itemize}

\begin{table}[h]
\caption{Per-seed test metrics for all $7$ tasks $\times$ $5$ methods $\times$ $3$ seeds. Seeds are anonymised as Run~1/2/3.}
\label{tab:per_seed}
\centering
\footnotesize
\setlength{\tabcolsep}{6pt}
\begin{tabular}{llccccc}
\toprule
Task & Run & VML & VML$\times$5 ens & VML$\times$5 best & VPP-MH & VPP-SMC \\
\midrule
\multirow{3}{*}{Linear (MSE $\downarrow$)}
 & 1 & $16.80$ & $10.45$ & $14.48$ & $0.47$ & $0.76$ \\
 & 2 & $6.99$  & $2.57$  & $2.06$  & $0.69$ & $4.13$ \\
 & 3 & $2.24$  & $0.61$  & $1.08$  & $1.86$ & $2.41$ \\
\midrule
\multirow{3}{*}{Poly (MSE $\downarrow$)}
 & 1 & $68.05$ & $1.97$  & $4.41$ & $7.59$  & $4.64$ \\
 & 2 & $91.92$ & $73.91$ & $7.41$ & $10.24$ & $1.85$ \\
 & 3 & $64.34$ & $9.06$  & $1.52$ & $9.22$  & $2.81$ \\
\midrule
\multirow{3}{*}{Sine (MSE $\downarrow$)}
 & 1 & $14.46$ & $5.51$ & $1.85$ & $0.71$ & $0.21$ \\
 & 2 & $0.19$  & $1.46$ & $0.15$ & $1.18$ & $0.10$ \\
 & 3 & $0.41$  & $0.79$ & $0.36$ & $0.91$ & $0.87$ \\
\midrule
\multirow{3}{*}{Two Blobs (acc.\ $\uparrow$)}
 & 1 & $95.0$ & $97.5$  & $97.5$  & $95.0$  & $95.0$  \\
 & 2 & $95.0$ & $100.0$ & $100.0$ & $100.0$ & $100.0$ \\
 & 3 & $50.0$ & $100.0$ & $100.0$ & $100.0$ & $100.0$ \\
\midrule
\multirow{3}{*}{Two Circles (acc.\ $\uparrow$)}
 & 1 & $100.0$ & $100.0$ & $100.0$ & $100.0$ & $100.0$ \\
 & 2 & $92.5$  & $95.0$  & $97.5$  & $100.0$ & $100.0$ \\
 & 3 & $100.0$ & $100.0$ & $100.0$ & $100.0$ & $100.0$ \\
\midrule
\multirow{3}{*}{Sum Parity (acc.\ $\uparrow$)}
 & 1 & $96.7$  & $80.0$ & $100.0$ & $90.0$  & $100.0$ \\
 & 2 & $100.0$ & $61.7$ & $100.0$ & $81.7$  & $100.0$ \\
 & 3 & $65.0$  & $60.0$ & $100.0$ & $100.0$ & $100.0$ \\
\midrule
\multirow{3}{*}{Contains Zero (acc.\ $\uparrow$)}
 & 1 & $36.7$  & $100.0$ & $100.0$ & $100.0$ & $100.0$ \\
 & 2 & $100.0$ & $73.3$  & $100.0$ & $100.0$ & $100.0$ \\
 & 3 & $98.3$  & $100.0$ & $100.0$ & $100.0$ & $100.0$ \\
\bottomrule
\end{tabular}
\end{table}

\clearpage

\section{Verbatim Particle Examples}
\label{app:particles}

The interpretability claim of Section~\ref{sec:method}---each particle is a complete, human-readable hypothesis---is best illustrated by quoting the actual text VPP recovers. The examples below are top-weighted particles from VPP-SMC ($K{=}10$) on three tasks where the method achieves $100\%$ test accuracy. The text is reproduced verbatim from the saved particle files; only mathematical notation has been re-typeset.

\paragraph{Sum parity (test accuracy $100\%$, top-particle weight $0.111$).}
\begin{quote}
\textit{``The input is a sequence of $4$ integers. You need to output the class label: an integer in the set \{0, 1\}. The class label is $1$ if the sum of the integers in the sequence is even, and $0$ otherwise.''}
\end{quote}
The posterior collapses to $4$ unique paraphrases of this rule across $10$ particles, each with the same likelihood; the BMA prediction inherits perfect accuracy from any of them. The four paraphrases differ only in surface form (whether ``sum is even'' is restated as ``digit sum is even,'' whether the variable name appears, etc.).

\paragraph{Two circles (test accuracy $100\%$, top-particle weight $0.125$).}
\begin{quote}
\textit{``You are designed to do binary classification. The input is a 2D point $(x_1, x_2)$; you need to output the class label, i.e., an integer in the set \{0, 1\}. The decision boundary is a circle centered at the origin $(0,0)$ with a radius of approximately $0.615$. If the distance from the origin to the input point, calculated as $\sqrt{x_1^2 + x_2^2}$, is less than or equal to the radius, output $1$; otherwise, output $0$.''}
\end{quote}
The numerical radius the particle settles on ($0.615$ here, $0.53$ in another high-weight particle) lies inside the true decision boundary; the ensemble averages out the residual uncertainty over the exact radius. Across the $10$ surviving particles, the radius estimates span $0.45$--$0.69$, all of which classify the test set perfectly because the geometry of the task is generous to slight under-estimates of the radius.

\paragraph{Contains zero (test accuracy $100\%$, top-particle weight $0.111$).}
\begin{quote}
\textit{``The model takes a sequence of $4$ integers as input and outputs a class label: $1$ if the sequence contains at least one $0$; $0$ if the sequence does not contain any $0$s.''}
\end{quote}
The hypothesis converges quickly (within the first few SMC steps) and remains stable across mutations.

\paragraph{Failure-mode comparison.}
On the same dataset (contains-zero), single-run VML ends with the following hypothesis at $36.7\%$ test accuracy:
\begin{quote}
\textit{``\ldots\ The model takes a sequence of $4$ integers as input and outputs a class label, which is $1$ if the sequence contains exactly two identical integers that are not adjacent and are separated by at least one different integer, and $0$ otherwise.''}
\end{quote}
This is a structurally plausible but completely wrong rule---it is the kind of pattern a careful human might propose after looking at a small number of examples and over-fitting to the surface structure of those examples. VML has no mechanism to recover from such a hypothesis once the optimizer commits to it: every subsequent revision starts from this incorrect rule and either preserves the error or, occasionally, drifts in unrelated directions. VPP-MH and VPP-SMC on the same dataset both reach $100\%$ accuracy because at least one particle finds the correct rule, and the accept/reject step (VPP-MH) or likelihood reweighting (VPP-SMC) prevents the wrong rule from dominating the BMA prediction.

\clearpage

\section{Sequential Monte Carlo Dynamics}
\label{app:smc_dynamics}

VPP-SMC's training trajectory is non-monotonic: rather than smoothly converging to a single posterior mode, the effective sample size oscillates as particles are reweighted, resampled, and mutated. Table~\ref{tab:ess} reports the per-step ESS on six tasks for a single representative seed, with the threshold $K/2 = 5$ marking when resampling is triggered.

Two patterns emerge. On classification and pattern-discovery tasks, the ESS rapidly stabilises near the maximum value of $K = 10$ once the population has identified the correct rule (sum parity from step $7$ onward, contains-zero from step $5$). On regression tasks, the ESS oscillates throughout training because no single arithmetic hypothesis dominates: the population continues to mix between roughly equally good arithmetic refinements until the end. The minimum ESS is reached at different steps for different tasks (sum parity at step $3$, contains-zero at step $4$, polynomial regression at the final step), which the resampling-on-threshold mechanism handles automatically by triggering systematic resampling whenever the threshold is crossed.

\begin{table}[h]
\caption{Effective sample size of VPP-SMC over the $20$ training steps for six tasks (single representative seed). The ESS threshold for resampling is $K/2 = 5$.}
\label{tab:ess}
\centering
\footnotesize
\setlength{\tabcolsep}{4pt}
\begin{tabular}{rcccccc}
\toprule
Step & Sum parity & Two circles & Contains zero & Linear & Poly & Sine \\
\midrule
$0$  & $9.33$ & $3.13$ & $6.46$ & $1.66$ & $4.24$ & $7.15$ \\
$1$  & $6.76$ & $3.62$ & $6.01$ & $6.72$ & $5.19$ & $7.31$ \\
$2$  & $6.95$ & $9.72$ & $6.68$ & $4.28$ & $1.82$ & $6.28$ \\
$3$  & $1.20$ & $8.56$ & $7.06$ & $4.40$ & $6.08$ & $3.92$ \\
$4$  & $8.49$ & $9.22$ & $1.03$ & $7.18$ & $3.41$ & $6.61$ \\
$5$  & $5.43$ & $9.56$ & $10.0$ & $5.32$ & $5.32$ & $7.35$ \\
$6$  & $3.54$ & $9.80$ & $10.0$ & $3.54$ & $2.17$ & $7.50$ \\
$7$  & $9.00$ & $9.35$ & $10.0$ & $3.79$ & $4.50$ & $6.24$ \\
$10$ & $9.00$ & $6.81$ & $10.0$ & $3.85$ & $1.31$ & $3.95$ \\
$15$ & $9.00$ & $7.42$ & $9.00$ & $2.53$ & $2.38$ & $4.38$ \\
$19$ & $9.00$ & $8.00$ & $9.00$ & $1.65$ & $1.00$ & $2.78$ \\
\midrule
mean   & $7.93$ & $7.61$ & $8.46$ & $4.18$ & $3.12$ & $5.78$ \\
min    & $1.20$ & $3.13$ & $1.03$ & $1.62$ & $1.00$ & $2.78$ \\
\bottomrule
\end{tabular}
\end{table}

\paragraph{Diversity at convergence.} The number of unique verbal hypotheses surviving in the final particle set varies with task difficulty and across seeds. On sum parity, the population converges to $4$--$9$ unique paraphrases of the correct rule (typically a small number on simpler seeds, more on harder ones); on contains-zero, $5$--$8$ unique hypotheses survive; on the regression tasks, $8$--$10$ unique hypotheses are present at the end, reflecting a more multimodal posterior over arithmetic rules of comparable likelihood. Diversity collapse is therefore not necessary for high accuracy: the posterior reaches the right answer either by concentrating on a single rule (for tasks with a clean symbolic structure) or by averaging over an ensemble of equivalently good ones (for tasks where the LLM's arithmetic limitations make exact convergence impossible).

\clearpage

\section{Per-Task Variance of Single VML Runs}
\label{app:vml_x5_full}

Table~\ref{tab:robustness} in the main paper showed the five individual VML runs that constitute VML$\times$5 on a representative dataset for three tasks. Table~\ref{tab:vml_x5_full} extends this to all seven tasks and all three seeds, exposing the run-to-run variance across the full benchmark suite. The pattern is consistent: regression tasks exhibit large per-run variance (factor of $\sim\!10$--$30\times$ between best and worst individual VML run on the same dataset), and classification and rule-discovery tasks show frequent catastrophic failures even when most runs succeed (e.g., a $0\%$-accuracy individual VML run on sum parity, multiple runs at $\le 50\%$ for sum parity on another seed, and two runs at $32\%$ for contains-zero). Across all $7\times 3 = 21$ task-seed combinations, the unweighted VML$\times$5 ensemble strictly underperforms the oracle best-of-$5$ on $11$ and ties on another $7$; only on three regression cases does prediction-averaging beat the single best run. VPP-MH and VPP-SMC (Table~\ref{tab:per_seed}) consistently land at the upper end of the spread.

\begin{table}[h]
\caption{Individual test metrics of the $5$ independent VML runs that constitute VML$\times$5, for all $7$ tasks $\times$ $3$ seeds (anonymised as Run~1/2/3). Format: $5$ comma-separated values per row. Lower is better for MSE (regression); higher for accuracy (classification, pattern discovery).}
\label{tab:vml_x5_full}
\centering
\footnotesize
\setlength{\tabcolsep}{6pt}
\begin{tabular}{llc}
\toprule
Task & Run & Individual runs \\
\midrule
\multirow{3}{*}{Linear (MSE $\downarrow$)}
 & 1 & $46.58,\ 38.16,\ 14.48,\ 27.45,\ 48.30$ \\
 & 2 & $5.50,\ 7.43,\ 36.44,\ 6.11,\ 2.06$ \\
 & 3 & $4.03,\ 1.08,\ 9.10,\ 27.06,\ 2.81$ \\
\midrule
\multirow{3}{*}{Poly (MSE $\downarrow$)}
 & 1 & $8.12,\ 16.01,\ 6.31,\ 4.41,\ 5.81$ \\
 & 2 & $115.22,\ 420.75,\ 110.29,\ 19.61,\ 7.41$ \\
 & 3 & $325.06,\ 17.55,\ 2.92,\ 1.52,\ 6.15$ \\
\midrule
\multirow{3}{*}{Sine (MSE $\downarrow$)}
 & 1 & $14.46,\ 22.91,\ 2.27,\ 16.24,\ 1.85$ \\
 & 2 & $11.95,\ 1.63,\ 1.19,\ 0.15,\ 0.74$ \\
 & 3 & $0.36,\ 0.57,\ 2.99,\ 5.05,\ 0.96$ \\
\midrule
\multirow{3}{*}{Two Blobs (acc.\ $\uparrow$)}
 & 1 & $97.5,\ 95.0,\ 72.5,\ 97.5,\ 97.5$ \\
 & 2 & $95.0,\ 72.5,\ 100,\ 100,\ 95.0$ \\
 & 3 & $100,\ 100,\ 70.0,\ 100,\ 100$ \\
\midrule
\multirow{3}{*}{Two Circles (acc.\ $\uparrow$)}
 & 1 & $97.5,\ 97.5,\ 100,\ 95.0,\ 67.5$ \\
 & 2 & $50.0,\ 97.5,\ 97.5,\ 90.0,\ 87.5$ \\
 & 3 & $100,\ 100,\ 97.5,\ 95.0,\ 100$ \\
\midrule
\multirow{3}{*}{Sum Parity (acc.\ $\uparrow$)}
 & 1 & $51.7,\ 48.3,\ 100,\ 41.7,\ 100$ \\
 & 2 & $100,\ 51.7,\ 50.0,\ 51.7,\ 48.3$ \\
 & 3 & $100,\ 48.3,\ 0.0,\ 45.0,\ 61.7$ \\
\midrule
\multirow{3}{*}{Contains Zero (acc.\ $\uparrow$)}
 & 1 & $100,\ 86.7,\ 61.7,\ 75.0,\ 100$ \\
 & 2 & $46.4,\ 53.3,\ 100,\ 51.7,\ 100$ \\
 & 3 & $31.7,\ 100,\ 100,\ 31.7,\ 100$ \\
\bottomrule
\end{tabular}
\end{table}


\end{document}